\pdfoutput=1
\documentclass{article}
\usepackage{spconf,amsmath,epsfig,cite}

\title{FGSD: A DATASET FOR FINE-GRAINED SHIP DETECTION IN HIGH RESOLUTION SATELLITE IMAGES}
\name{Kaiyan Chen, Ming Wu, Jiaming Liu, Chuang Zhang}
\address{Pattern Recognition and Intelligent System Lab, Beijing\\
University of Posts and Telecommunications, Beijing, China}
\begin{document}
%
\maketitle
\begin{abstract}
  Ship detection using high-resolution remote sensing images is an important task, which contribute to sea surface regulation. The complex background and special visual angle make ship detection relies in high quality datasets to a certain extent. However, there is few works on giving both precise classification and accurate location of ships in existing ship detection datasets. To further promote the research of ship detection, we introduced a new fine-grained ship detection datasets, which is named as FGSD. The dataset collects high-resolution remote sensing images that containing ship samples from multiple large ports around the world. Ship samples were fine categorized and annotated with both horizontal and rotating bounding boxes. To further detailed the information of the dataset, we put forward a new representation method of ships’ orientation. For future research, the dock as a new class was annotated in the dataset. Besides, rich information of images were provided in FGSD, including the source port, resolution and corresponding GoogleEarth's resolution level of each image. As far as we know, FGSD is the most comprehensive ship detection dataset currently and it'll be available soon. Some baselines for FGSD are also provided in this paper. You are also welcome to contact us at  ckyan@bupt.edu.cn to get the dataset.
\end{abstract}
\section{Introduction}
\label{sec:intro}
\begin{figure}[htb]
  \centering
  \centerline{\epsfig{figure=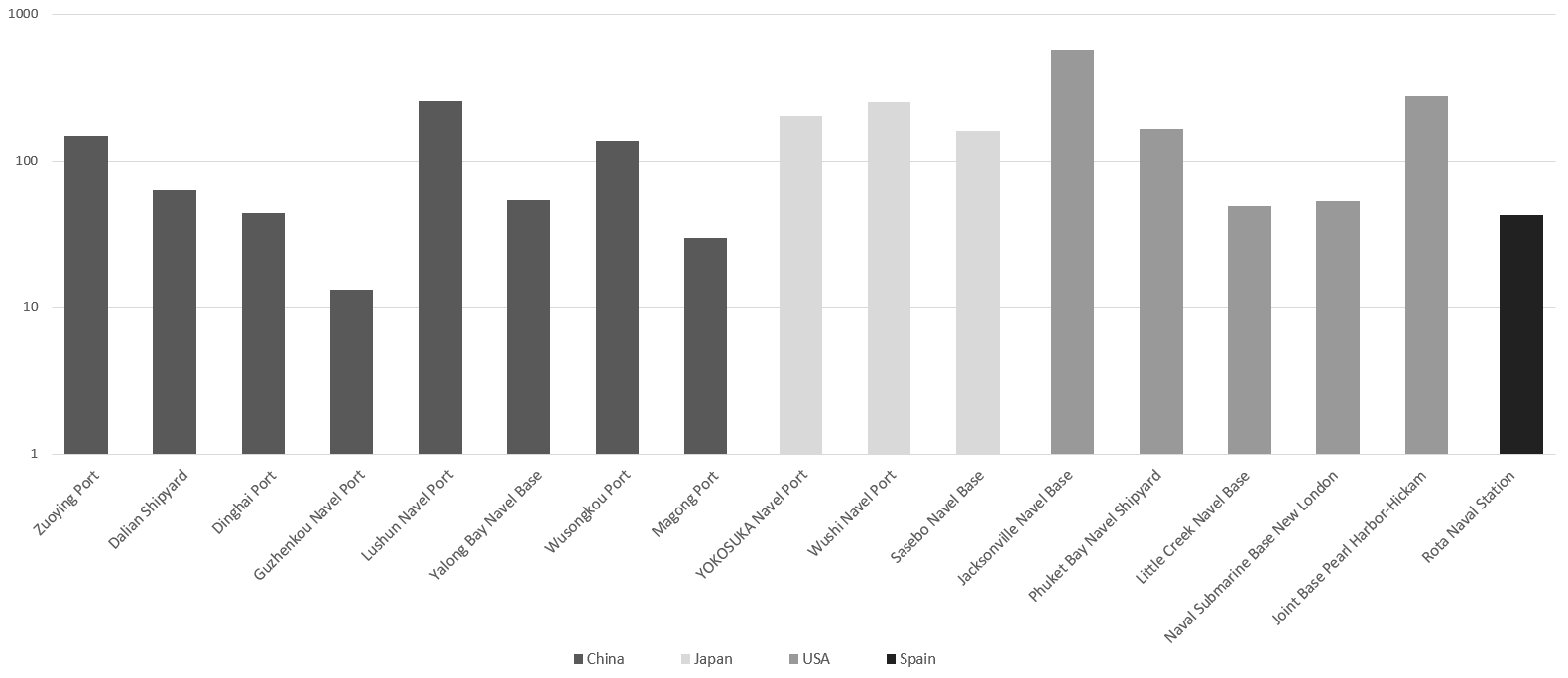,width=8.5cm}}
  \caption{Distribution of images quantities of every port. Ports of different countries are in different colors.}
  \label{fig:fig1}
\end{figure}
Ship detection based on remote sensing images, which aims to classified and locate ship samples using bounding boxes, has always been an important task. With the development of remote sensing technology in recent years, high-resolution remote sensing images make it possible to detect and classify ships more accurately. Ship detection, especially for near shore ships, plays an important supporting role in maritime management and trade estimates and so on. However, unlike target detection in natural scenes, remote sensing images always have large width, complex backgrounds, and multi-directional orientation, which requires the algorithm to be more robust and a high quantity dataset is also needed.

Remote sensing images that often used in ship detection task include high-resolution Synthetic Aperture Radar (SAR) images and high-resolution remote sensing optical  images. SAR image-based ship detection datasets generally do not classify ships in specific categories because of SAR images’ lack of detail features.

There are also some ship detection dataset  based on high-resolution remote sensing optical  images, such as DOTA\cite{Xia2018DOTA} ,NWPU VHR-10\cite{Cheng2014Multi}, but most of these dataset group all ships into one category. The most used dataset for ship detection in remote sensing optical  images currently is HRSC2016\cite{Liu2017A}, which collects images from six famous ports and annotated 22 classes of ships, but the images in HRSC2016 only covers limited ports and the category of ships are limited too, there is still much room for improvement.

\begin{figure*}[htb]
  \centering
  \centerline{\epsfig{figure=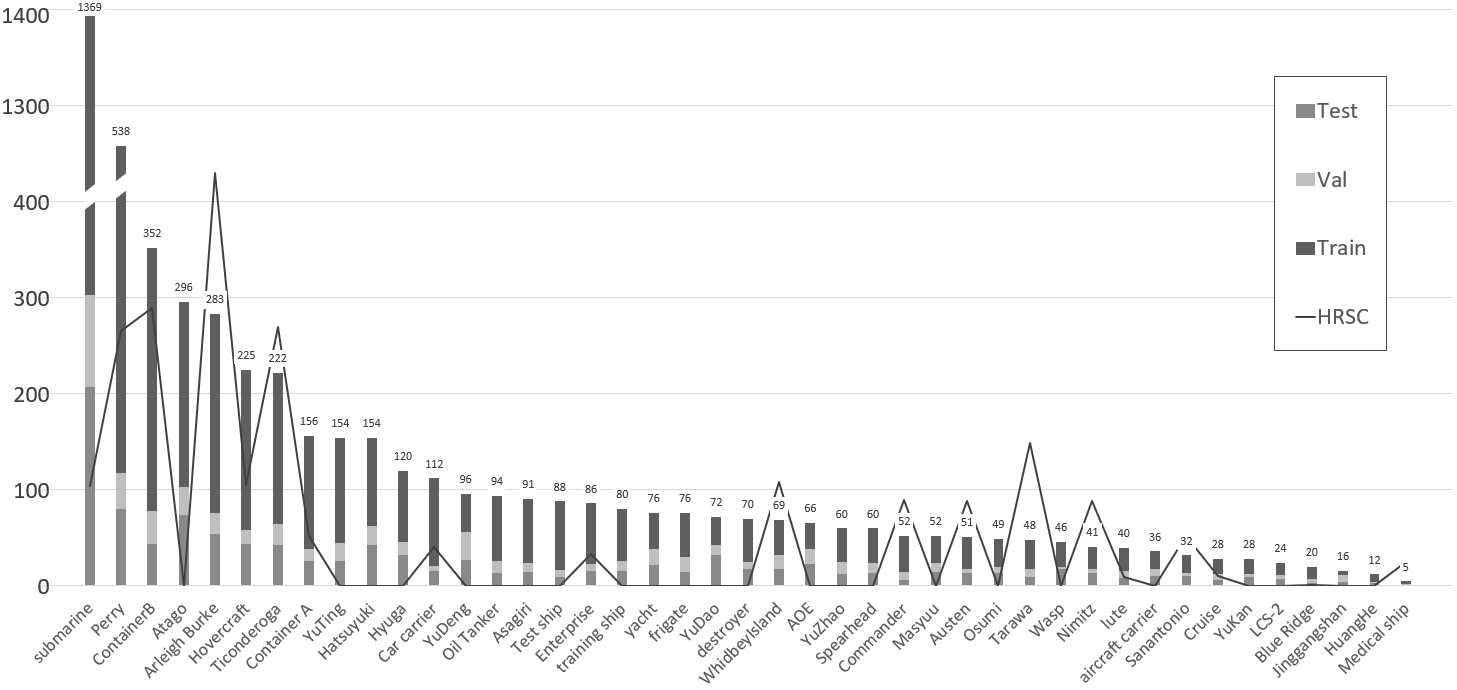,width=17.8cm,height=6.5cm}}
  \caption{The distribution of splited dataset and comparison with HRSC2016.}
  \label{fig:figure2}
\end{figure*}
In view of the above considerations, a larger and more comprehensive datasets for ship detection is introduced in this paper. Images from 17 large ports in four countries were collected and 43 classes of ships were labeled. Then different categorization methods were applied in all these 43 classes ships, further adding multilevel label for each instances. Except for ships, the dock as a new class were also annotated in FGSD for probable future research. All the ship samples in the dataset were annotated by both  common bounding box and rotated bounding boxes. Considering the orientation of ships, we also present a new concise representation of the ship's orientation. Besides, detailed information including source port's ID, resolution and corresponding GoogleEarth's resolution level are annotated in FGSD. As far as we know, FGSD is the largest fine ship detection dataset. We also benchmark some mainstream ship detection methods in FGSD as our baseline in this paper for further development.
\section{Setup Dataset}
\label{sec:format}
\subsection{Image collection}
\label{ssec:subhead}
We choose to collect images in google earth platform in several resolution levels including historical images. In order to ensure the  comprehensiveness of the dataset samples, we collected ship samples from 17 large ports including China, Japan, the United States and spain. Fig.\ref{fig:fig1} shows the quantities of images of every port.Taking into account the large differences in size between different types of ships, we performed image collection at multiple resolution levels. The pixel resolution of images in FGSD ranged from 0.12m to 1.93m. To ensure the data volume of each category of samples, map  shifts and spins were performed while sampling some ship samples, which greatly increased the data volume and ensure the diversity of the dataset.

After images collection, we got a total of 4736 pictures, In order to keep a balance between size of ships in images and the quanlity of images, the size of images were unified to $930\times 930$ in FGSD. 

\subsection{Categorization}
\label{ssec:subhead}
By data cleaning and samples labeling, 2612 images were selected and labeled from collected images, including 5634 labeled ship samples. After train-test split, we got a total of 1917 images including 3964 ship samples for training set, a total of 268 images including 590 ship samples for verification set and a total of 427 images including 1080 ship samples of test set, the sample distribution of each category is shown in Fig.\ref{fig:figure2}. Besides, a new class, dock, is annotated with rotate bounding box for probable future works, and a total of 1366 docks were annotated in the dataset. We compared FGSD with HRSC2016\cite{Liu2017A} in categories as is shown in Fig.\ref{fig:figure2}, it can be seen that we have far more categories and samples than HRSC2016.

\begin{figure}[htb]
  \centering
  \centerline{\epsfig{figure=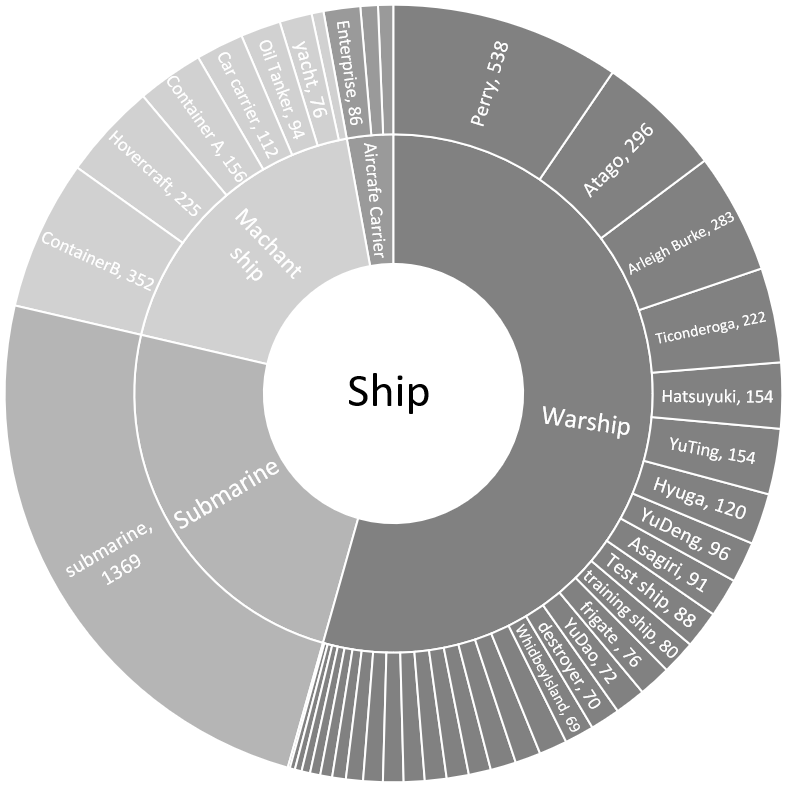,width=5cm}}
  \caption{Multilevel lable of ships and corresponding quantities.}
  \label{fig:figure3}
\end{figure}

We carefully choose and labeled 43 categories of ships and a 'dock' category in the dataset. All these ships were further labeled with multi-level labels follows\cite{Liu2017A} to some extent. The 43 classes of ships were divided into 4 Level-2 categories, including warship, carrier, submarine and civil ship. And all ships shares the same level-1 label ‘ship’. The multilevel label and corresponding distribution is shown in Fig.\ref{fig:figure3}.

As is mentioned above, some classes of ships only differ slightly while others differ a lot. Those categories that have slightly differences can be seen as fine-grained subsets. So we further labeled those fine-grained categories, including Aircraft Carriers, Warships, Amphibious assault ship and Supply ship. Fig.\ref{fig:figure5} shows some examples of subsets AircraftCarriers and Warships.
\begin{figure}[htb]
  \centering
  \centerline{\epsfig{figure=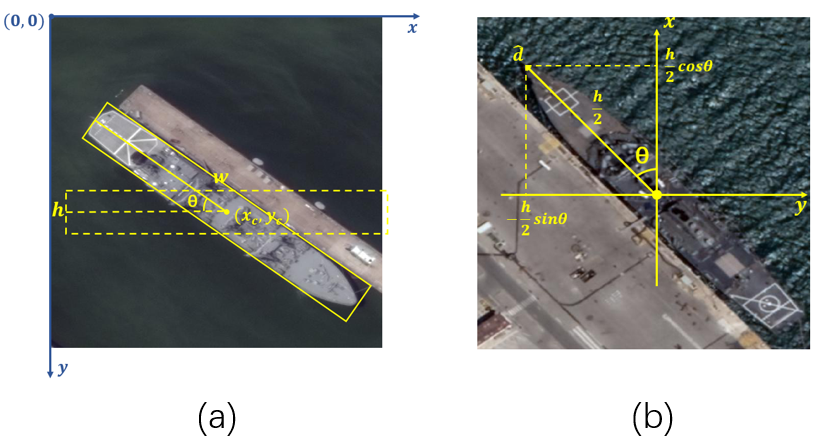,width=8cm}}
  \caption{Annotation methods of oriented bounding box. $(a)$ annotation method of a rotated bounding box; $(b)$ annotation method of orientation}
  \label{fig:figure4}
\end{figure}
\subsection{Annotation}
\label{ssec:subhead}
\begin{figure}[htb]
  \centering
  \centerline{\epsfig{figure=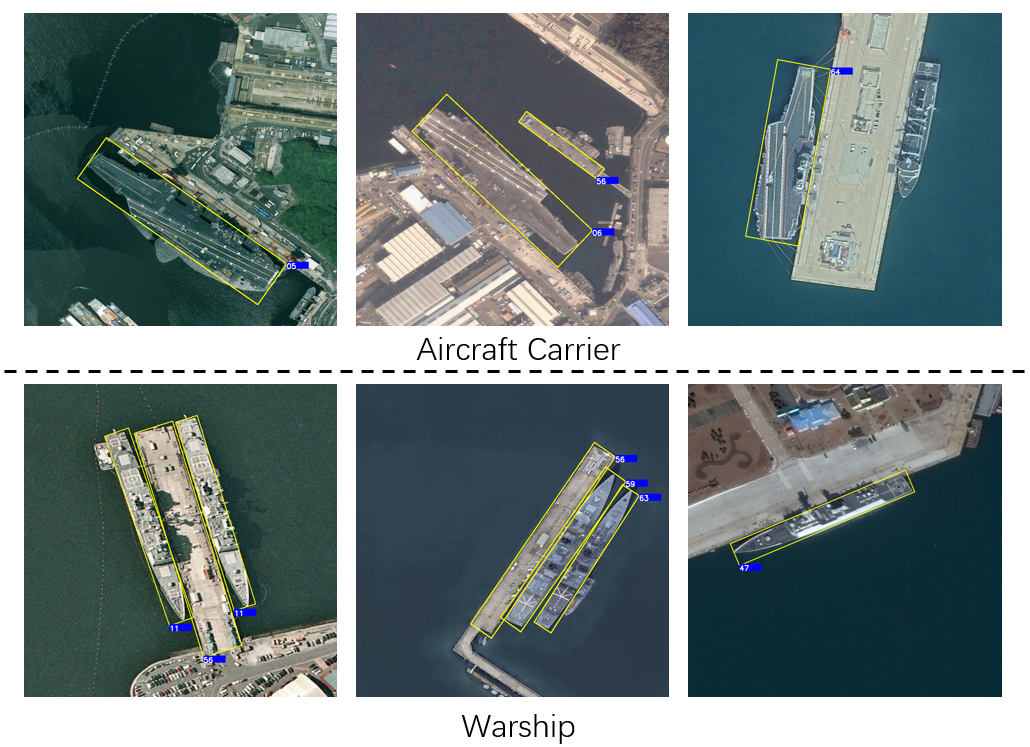,width=7cm}}
  \caption{Examples of labeled images and fine-grained subsets.}
  \label{fig:figure5}
\end{figure}
Different from object detection in natural scenes, remote sensing images have a unique top-down perspective, which makes the targets in the remote sensing images have different orientations. When the ships in the image are densely arranged, locating ships using horizontal rectangular bounding boxes will cause a lot of missed detection, and the horizontal rectangular bounding box will also introduce a lot of background noise in remote sensing images. Therefore, ship samples in FGSD were annotated with both common used bounding box and rotated bounding box. Noticing that the rotated bounding box is annotated as $(x_{c},y_{c},w,h,\theta)$ \cite{Liu2017A} in FGSD, where $(x_{c},y_{c})$ denotes the center point of ship, and w, h denotes the width and height of standard bounding box, $\theta$ denotes the angle between standard box and rotated bounding box. Fig.\ref{fig:figure4} $(a)$ shows the annotation method for rotated bounding box.

To specify the annotation for ships, we also provide annotation for the ‘head’ of ships in a new way. In previous works, The HRSC2016\cite{Liu2017A} indicates the ship's orientation by marking the V-shaped point of the ship, and DOTA gives the orientation by marking the starting corner point of the orientate bounding box. We put forward a more intuitive and concise representation of the orientation of ships. As is shown in the Fig.\ref{fig:figure4} $(b)$, given a target with orientation bounding box $(x_{c},y_{c},w,h,\theta)$, create coordinate axis with the center point of the target as the origin. The orientation of a ship can be expressed as a vector $\hat{d}=(\hat{x},\hat{y})$ , which points from the center point of ship to the head of ship. And the orientation can be annotated by $(d_{1},d_{2})$, where $d_{1},d_{2}\in\left \{ -1,0,1 \right \}$ indicates the positive and negative direction of $\hat{x}$ and $\hat{y}$ respectively. $\hat{x}$ and $\hat{y}$ is calculated by:
\[\hat{x}=d_{1}\cdot\frac{max(w,h)}{2}sin\theta,\; \; \hat{y}=d_{2}\cdot\frac{max(w,h)}{2}cos\theta\]

Noticing that this annotation method can also be applied to the orientation representation of any other rotating target.

Besides, in the dataset's pubic version, we provide details information of each images in the annotation file, including the source-port's ID of the image, the resolution and corresponding GoogleEarth's resolution level of the image. Some labeled examples are given in Fig.\ref{fig:figure5}.

\section{BASELINE}
\label{sec:pagestyle}

\begin{table*}[]
  \caption{Evaluation results AP for each category. Baseline R2CNN for rotated bounding box detection and Faster-RCNN for horizontal bounding box detecion were denoted as $BL_R$ and $BL_H$ respectively. And the mAP of $BL_R$ is 0.65, the mAP of $BL_H$ is 0.7.}
  \label{tab:Table1}
  \centering
  \begin{tabular}{|c|cc|c|cc|c|cc|c|cc|}
  \hline
  CLS    & $BL_R$ & $BL_H$ & CLS    & $BL_R$ & $BL_H$ & CLS    & $BL_R$ & $BL_H$ & CLS    & $BL_R$ & $BL_H$ \\ \hline
  Nim.   & 0.63  & 0.67  & Hov.   & 0.46  & 0.5   & YuD.   & 0.42  & 0.46  & Spea.  & 0.77  & 0.79  \\
  Ent.   & 0.19  & 0.33  & yacht  & 0.33  & 0.33  & YuT.   & 0.6   & 0.43  & Hat.   & 0.58  & 0.73  \\
  Arl.   & 0.75  & 0.8   & Cont.B & 0.59  & 0.62  & YuK.   & 0.61  & 0.44  & Mas.   & 0.87  & 0.87  \\
  Whi.   & 0.39  & 0.65  & Cru.   & 0.5   & 0.5   & YuZ.   & 0.91  & 0.95  & Ata.   & 0.78  & 0.8   \\
  Per.   & 0.79  & 0.86  & Sub.   & 0.6   & 0.65  & J.G.S. & 0.56  & 0.5   & dest.C & 0.59  & 0.74  \\
  San.   & 0.78  & 0.73  & lute   & 0.75  & 0.75  & Hua.H. & 0.58  & 0.67  & frig.C & 0.65  & 0.84  \\
  Tic.   & 0.86  & 0.88  & Med.   & 0.81  & 0.74  & YuD.   & 0.47  & 0.47  & AOE    & 0.71  & 0.83  \\
  Aus.   & 0.46  & 0.5   & Car.   & 0.88  & 0.88  & Test.  & 0.45  & 0.65  & Asa.   & 0.54  & 0.6   \\
  Tar.   & 0.78  & 0.87  & Osu.   & 0.88  & 0.89  & Tra.   & 0.56  & 0.78  & Air.C  & 0.9   & 0.93  \\
  Cont.A & 0.39  & 0.39  & Wasp   & 0.78  & 0.85  & Oil.T  & 0.81  & 0.88  &        &       &       \\
  Com.   & 0.68  & 0.68  & Hyu.   & 0.94  & 0.94  & LCS-2  & 0.85  & 0.85  &        &       &       \\ \hline
  \end{tabular}
  \end{table*}
There are two tasks in ship detection: the common bounding box detection and rotating bounding box detection. Both tasks need to predict the bounding box and classification for each ship samples in images. The common used bounding box and rotated bounding box are required in two tasks respectively.

Lot of works on algorithms for ship detection have been done in previous. Early ship detection methods such as \cite{2010ITGRS..48.3446Z} were mostly based on manually designed feature operators. \cite{Yuan2017Ship,rs11050531} used convolution network to build ship detection algorithms. Many algorithms for orientation objects detection\cite{JiangR2CNN,YangSCRDet} can also been transferred to ship detection task.

We choose the Faster-RCNN\cite{RenFaster} framework and R2CNN\cite{JiangR2CNN} as our baselines for horizontal bounding box detection and rotated bounding box detection respectively, the evaluation results is shown in Table\ref{tab:Table1}. For fare comparison, ResNet-101 is used as the backbone in above mentioned baselines. The criteria average precision(AP) and mean average precision(mAP) are used for evaluation.

\section{CONCLUSION}
\label{sec:typestyle}
For the purpose of promoting research on ship detection, a new dataset, named FGSD, for fine-grained ship detection in high resolution satellite images was introduced in this paper. As far as we know, FGSD is the largest fine-grained ship recognition dataset in remote sensing fields. Rich information has been added in the dataset, including multi-level label, multiple labeling methods and detailed classification for ships. And a new description methods for orientation of targets was proposed for further works. By analyzing some well performed algorithm on FGSD, baselines results were given in this paper. 

Further works will be done to extent FGSD, and future work, such as  combine ship detection algorithms with fine-grained classification algorithms to improve the performance, will be done to promote the development of ship recognition fields too.
\label{sec:ref}

\bibliographystyle{IEEEbib}
\bibliography{strings,refs}

\end{document}